\documentclass[%
 reprint,
 superscriptaddress,
 preprintnumbers,
 nofootinbib,
 amsmath,amssymb,
 aps,
 longbibliography,
]{revtex4-2}

\usepackage{graphicx}
\usepackage{dcolumn}
\usepackage{bm}
\usepackage{xcolor}
\usepackage{array}
\usepackage{booktabs}
\usepackage{hyperref}
\usepackage{float}
\usepackage{enumitem}
\usepackage{tikz}
\usetikzlibrary{arrows.meta, positioning}
\usepackage{pgfplots}
\pgfplotsset{compat=1.18}
\usepgfplotslibrary{groupplots}
\definecolor{aichatgpt}{HTML}{9ECAE1}
\definecolor{aiclaude}{HTML}{4292C6}
\definecolor{aideepseek}{HTML}{08519C}
\definecolor{humanpink}{RGB}{225,85,136}

\newsavebox{\propbox}
\newenvironment{proposalbox}[1]
 {\par\medskip\begin{widetext}\setlength{\fboxsep}{10pt}%
 \begin{lrbox}{\propbox}%
 \begin{minipage}{\dimexpr\textwidth-2\fboxsep-2\fboxrule\relax}%
 \hfill\textbf{\large #1}\par\medskip}
 {\end{minipage}\end{lrbox}%
 \noindent\fbox{\usebox{\propbox}}\end{widetext}\medskip}

\begin{document}
 \preprint{IPMU26-0030}
\title{AI's Capability in Assisting Scientific Research in Physics, Astrophysics, and Cosmology II: Project Planning and Proposal Evaluation}

\author{Jia Liu$^{1,2}$}
\email{jia.liu@ipmu.jp}
\noaffiliation
\author{Veena Krishnaraj$^{3}$}
\noaffiliation
\author{Kateryna Vovk$^{1,2}$}
\noaffiliation
\author{Kosuke Aizawa$^{4,1,2}$}
\noaffiliation
\author{Adrian E.~Bayer$^{5,3}$}
\noaffiliation
\author{Linda Blot$^{1,2}$}
\noaffiliation
\author{Jessica Cowell$^{6,1,2}$}
\noaffiliation
\author{Suyog Garg$^{7}$}
\noaffiliation
\author{Jonathan Gr\'ee$^{8}$}
\noaffiliation
\author{Anamaria Hell$^{1,2}$}
\noaffiliation
\author{Ben Horowitz$^{1,2}$}
\noaffiliation
\author{Masaya Ichikawa$^{9}$}
\noaffiliation
\author{Kanyuni Iemoto$^{10}$}
\noaffiliation
\author{Keigo Kondo$^{11,7}$}
\noaffiliation
\author{Zacharie Lorsin$^{8}$}
\noaffiliation
\author{Kevin McCarthy$^{1,2}$}
\noaffiliation
\author{Jamie Robinson$^{3}$}
\noaffiliation
\author{Miguel Ruiz-Granda$^{12,13}$}
\noaffiliation
\author{Leander Thiele$^{1,2}$}
\noaffiliation
\author{Ievgen Vovk$^{14}$}
\noaffiliation
\author{Mingshen Zhou$^{15,16}$}
\noaffiliation
\collaboration{Author affiliations are listed at the end of the paper.}

\noaffiliation

\begin{abstract}
We investigate how well large language models (LLMs) can assist scientific project planning and proposal evaluation. One-page project plans were independently generated for eight expert-conceived research projects in physics, astrophysics, and cosmology by human researchers and three contemporary LLMs (ChatGPT, Claude, and DeepSeek; mid-2025 models, used with their default tool access). The resulting 32 proposals were blindly evaluated by four human reviewers and two newer frontier LLMs (Claude Opus 4.8 and ChatGPT Pro 5.5) using a four-aspect evaluation rubric. Reviewers were also asked to identify whether each proposal was written by a human or an AI. Human reviewers rated human- and AI-written proposals similarly overall, whereas both AI reviewers scored AI-written proposals about one point higher (on a five-point scale) than human-written proposals. Human reviewers correctly identified human- and AI-written proposals 72\% and 79\% of the time, respectively, while both AI reviewers correctly classified all 32 proposals (100\%). These results suggest that current LLMs can produce project plans comparable to human-written ones in the eyes of human reviewers, but that AI reviewers show a systematic preference for AI-generated proposals. Our results suggest caution when deploying LLMs widely in proposal preparation and evaluation.
\end{abstract}
\maketitle
\section{Introduction}
Artificial intelligence (AI), and in particular large language models (LLMs), is increasingly being used at every stage of the scientific process, from literature review and ideation to coding, analysis, and writing. Yet most assessments of AI capability rely on benchmarks whose answers are already known, such as standardized exams or curated questions, which measure how well a model recovers existing knowledge rather than how well it operates at the research frontier, where neither humans nor machines yet know the answer.
In physics, astrophysics, and cosmology, deep learning has been widely applied to accelerate computation and achieve more accurate parameter inferences (e.g., \cite{Hezaveh2017,VillaescusaNavarro2021}). More recently, the rise of LLMs has extended AI from numerical tasks to language- or reasoning-related ones \cite{Nguyen2023,Zhou2024,Mitchell2023,Gao2023,Lu2024,Shcherbiak2024,Panickssery2024,Si2024,Wataoka2024,Liang2024,Sadasivan2023,Ren2025,Sikimic2025,Sandstrom2026, Thorne2026, denario, Hell:2026dqx}.
Notably, \cite{Si2024} found that LLMs can generate research ideas that expert reviewers rate as more novel than human-generated ones; \cite{Liang2024} found that LLMs can provide feedback on papers that researchers often find useful; \cite{Shcherbiak2024} reported low inter-reviewer agreement together with quality-rating inflation by AI (also see \cite{Panickssery2024, Wataoka2024}). What remains comparatively unexplored is a controlled, blinded comparison of human and AI performance on an open-ended research-planning task, evaluated by both human and AI reviewers, with attention to evaluator bias. In a companion paper \cite{Paper1}, we examine AI's capability in literature review.

This work is motivated by three questions. First, how capable is AI at frontier-science project planning, in a regime where there is no known answer? Second, and more exploratory, can humans and AI identify whether a piece of scientific text was written by a human or by an AI, and what reasoning underlies those judgments? Third, how do humans and AI evaluate research plans, and do their evaluations carry systematic biases? The last question speaks directly to the future of funding agencies and to how researchers should prepare proposals in an era when both authorship and review may be AI-assisted.

To address these questions, we conducted a controlled study on eight expert-conceived research projects spanning physics, astrophysics, and cosmology. For each project, a human researcher and three AI assistants independently wrote a one-page proposal from the same title, background, and goal, which were provided verbatim as the common starting point. All 32 proposals were then assessed blindly by human reviewers and LLMs. Three findings stand out: (1) humans identified authorship correctly $\sim70\%$ of the time, whereas the most capable AI reviewers did so essentially perfectly; (2) human reviewers judged AI-written proposals to be no worse than expert-written ones; yet (3) the AI reviewers systematically scored AI-written proposals above human-written ones, a pro-AI bias absent from the human panel. These results carry concrete implications for AI-assisted proposal writing and for the design of review processes as AI is increasingly used in both proposal writing and review.

The remainder of this paper is organized as follows. Section~\ref{sec:methods} describes the methodology: the eight research projects, the four groups and their roles, the proposal-writing protocol, and the blind-evaluation procedure and rubric. Section~\ref{sec:results} presents the results on origin identification and on proposal quality, for both human and AI reviewers. Section~\ref{sec:conclusion} summarizes our findings and discusses their implications for AI-assisted research planning and proposal review.
\section{Methods}
\label{sec:methods}

This work addresses the project-planning stage of the scientific workflow: drafting a research proposal that specifies the scientific goal, the methodology, and the resources and timeline required to carry a project through to completion. To this end, we assembled a controlled corpus of proposals for eight expert-conceived research projects, each prepared independently by a human expert and by three AI assistants, under an identical template, prompt, and rubric, and collected blind quality ratings and human-versus-AI origin judgments from both human and AI reviewers.

The study is organized around four groups: human project planners (the experts), AI prompters, human reviewers, and AI reviewers. Several design choices were made to isolate proposal content from superficial cues: (1) every proposal followed an identical one-page template and section structure (Sec.~\ref{sec:proposal_writing}); (2) human-written proposals were passed through a light grammar-and-spelling pass to remove obvious human mistakes (Sec.~\ref{sec:proposal_writing}); and (3) all proposals were anonymized and uniformly formatted before review (Sec.~\ref{sec:evaluation}). The proposals were deliberately kept much shorter than a typical grant proposal (roughly one page) primarily to keep the workload manageable for the experts who designed the projects and for the reviewers who scored every submission. In total, the eight projects yielded $8\times4 = 32$ proposals (8 human, 24 AI), each scored by four human reviewers and two AI reviewers.

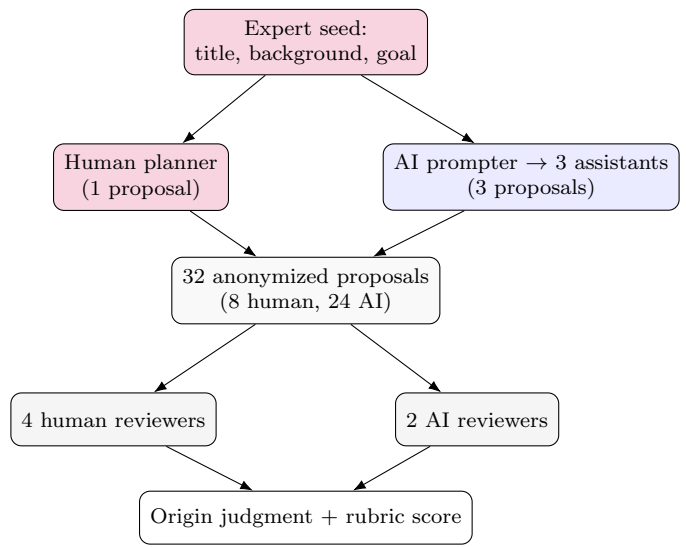
\begin{figure}[t]
\centering
\begin{tikzpicture}[
 font=\footnotesize, >=Latex,
 box/.style={draw, rounded corners, align=center, inner sep=4pt, minimum height=7mm},
 ai/.style={box, fill=blue!8},
 hu/.style={box, fill=humanpink!25},
 ev/.style={box, fill=gray!8},
]
\node[hu] (seed) {Expert seed:\\ title, background, goal};
\node[hu, below left=9mm and -6mm of seed] (hp) {Human planner\\ (1 proposal)};
\node[ai, below right=9mm and -6mm of seed] (am) {AI prompter $\to$ 3 assistants\\ (3 proposals)};
\node[box, fill=gray!5, below=24mm of seed] (corpus) {32 anonymized proposals\\ (8 human, 24 AI)};
\node[ev, below left=9mm and -6mm of corpus] (hr) {4 human reviewers};
\node[ev, below right=9mm and -6mm of corpus] (ar) {2 AI reviewers};
\node[box, below=22mm of corpus] (out) {Origin judgment $+$ rubric score};
\draw[->] (seed) -- (hp);
\draw[->] (seed) -- (am);
\draw[->] (hp) -- (corpus);
\draw[->] (am) -- (corpus);
\draw[->] (corpus) -- (hr);
\draw[->] (corpus) -- (ar);
\draw[->] (hr) -- (out);
\draw[->] (ar) -- (out);
\end{tikzpicture}
\caption{Overview of the study design. Each expert-provided seed (title, background, goal) is turned into one human-written and three AI-written one-page proposals; the 32 anonymized proposals are then assessed for authorship and quality by four human and two AI reviewers.}
\label{fig:workflow}
\end{figure}
\subsection{The eight research projects}
\label{sec:projects}
The eight projects span theoretical, computational, observational, and instrumentation work across cosmology, astrophysics, and physics. Each was conceived and outlined by a domain expert without any AI assistance; the expert provided a project title, a background paragraph, and a goal paragraph (Table~\ref{tab:projects} summarizes the set). These three elements were the common starting point given to both the human planner and the AI prompter for each project. The full titles, backgrounds, and goals for all eight projects are given in Appendix~\ref{app:project_background} and match those used in the companion Paper~I; Table~\ref{tab:projects} serves here as a quick reference.
\begin{table*}
\renewcommand{\arraystretch}{1.5}
\begin{tabular}{|l|l|l|}
\hline
\parbox[t]{1cm}{\raggedright \textbf{ID}} &
\parbox[t]{8.5cm}{\raggedright \textbf{Project title}} &
\parbox[t]{5cm}{\raggedright \textbf{Domain}} \\
\hline
\parbox[t]{1cm}{\raggedright AGN} &
\parbox[t]{8.5cm}{\raggedright MaNGA: AGN Duty Cycle} &
\parbox[t]{5cm}{\raggedright AGN / galaxy evolution} \\
\hline
\parbox[t]{1cm}{\raggedright LBG} &
\parbox[t]{8.5cm}{\raggedright The galaxy--dark matter halo connection of Lyman-break galaxies} &
\parbox[t]{5cm}{\raggedright Large-scale structure} \\
\hline
\parbox[t]{1cm}{\raggedright IA} &
\parbox[t]{8.5cm}{\raggedright Intrinsic alignments in varying environments} &
\parbox[t]{5cm}{\raggedright Weak lensing / intrinsic alignments} \\
\hline
\parbox[t]{1cm}{\raggedright AR} &
\parbox[t]{8.5cm}{\raggedright Prediction of debris emergence on laser-ablated sub-wavelength shapes for millimeter-wave anti-reflection} &
\parbox[t]{5cm}{\raggedright Instrumentation / laser fabrication} \\
\hline
\parbox[t]{1cm}{\raggedright RG} &
\parbox[t]{8.5cm}{\raggedright Radio Galaxies with HalfDome} &
\parbox[t]{5cm}{\raggedright CMB foregrounds / radio galaxies} \\
\hline
\parbox[t]{1cm}{\raggedright GW} &
\parbox[t]{8.5cm}{\raggedright Studying the environment of gravitational-wave black hole binaries with weak lensing maps} &
\parbox[t]{5cm}{\raggedright Gravitational waves / multi-messenger} \\
\hline
\parbox[t]{1cm}{\raggedright PTA} &
\parbox[t]{8.5cm}{\raggedright Forecasting the pulsar timing array sensitivity needed to measure deviations from general relativity} &
\parbox[t]{5cm}{\raggedright Gravitational waves / tests of GR} \\
\hline
\parbox[t]{1cm}{\raggedright SU2} &
\parbox[t]{8.5cm}{\raggedright Massive Yang--Mills theory (SU(2))} &
\parbox[t]{5cm}{\raggedright Theoretical cosmology / inflation} \\
\hline
\end{tabular}
\caption{The eight expert-conceived research projects used in this study. Titles are as written by the human experts. Each project was independently turned into one human-written and three AI-written one-page proposals.}
\label{tab:projects}
\end{table*}
\subsection{Participant roles}
\label{sec:roles}
The study involved four groups.

\textbf{Human project planners (experts).} For each of the eight projects, one human expert (graduate student or postdoc who works on that topic as part of their routine research) wrote the proposal without any AI assistance. The expert-provided title, background, and goal were the starting point for every proposal, both human and AI. To keep the human and AI proposals format-consistent, the experts followed a shared proposal template and instructions (Sec.~\ref{sec:proposal_writing}).

\textbf{AI prompters.} For each project, one AI prompter generated the AI proposals. Prompters were undergraduate or graduate students working outside the specific field of the project but within the broader area of physics, astrophysics, and cosmology. Each prompter took the expert-provided title, background, and goal and prompted three AI assistants with a fixed prompt template (Sec.~\ref{sec:proposal_writing}) to produce three AI proposals per project. Each proposal came from a single pass of the fixed prompt template, and the output was taken as-is, without iterative refinement.

\textbf{Human reviewers.} Four of this paper's authors served as blind reviewers. They are faculty members or senior postdocs who, while not necessarily experts on the specific topic of any given proposal, have sufficient knowledge to judge the projects, and as such mirror the composition of a typical grant review panel. They received the full set of anonymized proposals and, for each one, first judged whether it had been written by a human or an AI and then scored it against the predefined rubric (Sec.~\ref{sec:evaluation}).

\textbf{AI reviewers.} Two AI assistants were given the identical anonymized proposals and rubric and performed the same task as the human reviewers: an origin judgment (human vs.\ AI) followed by rubric scores (Sec.~\ref{sec:evaluation}). We used Claude Opus~4.8 (Anthropic) and ChatGPT Pro~5.5 (OpenAI), two of the more capable models commonly available to general researchers in mid-2026.

\subsection{Proposal preparation}
\label{sec:proposal_writing}
Every proposal, human or AI, followed the same one-page template with four headed sections: (1) Project Title, copied from the expert's title; (2) Background, a one-sentence summary; (3) Goal, a one-sentence summary (the expert's full background and goal were provided verbatim as input and here condensed to one sentence each, whereas the title was reproduced verbatim); and (4) Methodology, broken into no more than five major steps or phases (e.g., data preparation, modeling, analysis), each with an approximate completion time, in under 300 words. Proposals were to use a formal scientific tone appropriate for an academic review panel. This template is intentionally far shorter than a typical research or grant proposal, keeping the writing and review task tractable while still exercising the core planning skills of framing, methodology, and timeline. Whether the same patterns hold for full-length proposals is left to future work. The full instructions and prompt template are in Appendix~\ref{app:project_plan_prompt}.

The human proposals were written by the experts. Starting from the previously prepared background and goal paragraphs, each expert wrote a one-page plan following the template. To remove superficial stylistic and grammatical cues that might reveal human authorship, each human-written plan was then passed through ChatGPT (4o) with the instruction to correct typos and grammatical mistakes while making minimal changes to the content. This normalization preserved the scientific content while making the surface style more uniform across human and AI submissions.

The AI proposals were generated by the AI prompters between June and September 2025.  For each project, the prompter generated three proposals using three different assistants, ChatGPT (4o), Claude (Sonnet 4), and DeepSeek (V3), each prompted with the same template (Appendix~\ref{app:project_plan_prompt}) and with the expert's title, background, and goal. This yielded $8\times3 = 24$ AI proposals, which together with the 8 human proposals make up the corpus of 32.

\subsection{Proposal evaluation}
\label{sec:evaluation}
\begin{table*}[t]
\renewcommand{\arraystretch}{1.5}
\begin{tabular}{|l|l|l|l|l|l|}
\hline
\parbox[t]{2.3cm}{\raggedright \textbf{Aspect}} &
\parbox[t]{2.8cm}{\raggedright \textbf{5 (Excellent)}} &
\parbox[t]{2.8cm}{\raggedright \textbf{4 (Good)}} &
\parbox[t]{2.8cm}{\raggedright \textbf{3 (Adequate)}} &
\parbox[t]{2.8cm}{\raggedright \textbf{2 (Weak)}} &
\parbox[t]{2.8cm}{\raggedright \textbf{1 (Poor)}} \\
\hline
\parbox[t]{2.3cm}{\raggedright Clarity and Structure of Research Plan} &
\parbox[t]{2.8cm}{\raggedright Clear, logical sequence of steps; structured into coherent phases.} &
\parbox[t]{2.8cm}{\raggedright Mostly clear structure; small gaps in logic or detail.} &
\parbox[t]{2.8cm}{\raggedright Some structure present, but steps are vague or loosely connected.} &
\parbox[t]{2.8cm}{\raggedright Poorly structured or fragmented plan; lacks internal coherence.} &
\parbox[t]{2.8cm}{\raggedright No discernible structure; chaotic or missing entirely.} \\
\hline
\parbox[t]{2.3cm}{\raggedright Appropriateness of Methods to Scientific Goal} &
\parbox[t]{2.8cm}{\raggedright Methods are well-justified, state-of-the-art, and fit the scientific question.} &
\parbox[t]{2.8cm}{\raggedright Methods are suitable, though justification may be thin or options unexplored.} &
\parbox[t]{2.8cm}{\raggedright Methods are acceptable but basic or overly generic.} &
\parbox[t]{2.8cm}{\raggedright Methods are mismatched or poorly motivated.} &
\parbox[t]{2.8cm}{\raggedright Methods are incorrect, unworkable, or irrelevant.} \\
\hline
\parbox[t]{2.3cm}{\raggedright Resource and Tool Planning} &
\parbox[t]{2.8cm}{\raggedright Specifies required datasets, software, AI tools, and compute needs with precision.} &
\parbox[t]{2.8cm}{\raggedright Most tools and resources identified; some details vague.} &
\parbox[t]{2.8cm}{\raggedright Only basic resources mentioned; limited specificity.} &
\parbox[t]{2.8cm}{\raggedright Important resources missing or misunderstood.} &
\parbox[t]{2.8cm}{\raggedright No mention of data/tools; unrealistic resource assumptions.} \\
\hline
\parbox[t]{2.3cm}{\raggedright Feasibility, Timeline, and Risk Awareness} &
\parbox[t]{2.8cm}{\raggedright Timeline is realistic, well-paced; risks and contingencies are clearly addressed.} &
\parbox[t]{2.8cm}{\raggedright Feasible plan with some timing uncertainty; minor risk handling.} &
\parbox[t]{2.8cm}{\raggedright Rough timeline present but lacks detail or risk mitigation.} &
\parbox[t]{2.8cm}{\raggedright Unclear or unrealistic timeline; no mention of risks.} &
\parbox[t]{2.8cm}{\raggedright Infeasible timeline or no plan at all.} \\
\hline
\end{tabular}
\caption{Rubric used for the blind evaluation of the research proposals. Each proposal was scored from 1 to 5 on each of the four aspects by every reviewer (human and AI).}
\label{tab:project_plan_rubric}
\end{table*}
All 32 proposals were anonymized and standardized to a uniform visual format so that neither origin nor identity could be inferred from formatting. Each proposal was then independently assessed by four human reviewers and, separately, by two AI reviewers. For every proposal, each reviewer performed two tasks: (1) a binary judgment of whether the proposal was written by a human or by an AI; and (2) a quality rating on a predefined rubric.

The rubric scores each proposal from 1 (poor) to 5 (excellent) along four aspects: clarity and structure of the research plan; appropriateness of methods to the scientific goal; resource and tool planning; and feasibility, timeline, and risk awareness. The full scoring rubric, shown identically to both the human and AI reviewers, is given in Table~\ref{tab:project_plan_rubric}. 

The AI evaluation was performed later in the process, around mid 2026. We used two LLMs, Claude Opus~4.8 and ChatGPT Pro~5.5, each asked to score all 32 proposals on the rubric and to guess whether each proposal was human- or AI-written, using the same instructions provided to the human reviewers. The agreement between human and AI reviewers, both on origin classification and on rubric scores, is presented in Sec.~\ref{sec:results}.

\section{Results}
\label{sec:results}

We report two complementary analyses of the 32 proposals. First, we ask how reliably evaluators can identify whether a proposal was written by a human or an AI, and what cues drive those judgments (Sec.~\ref{sec:results_origin}). Second, we compare the quality scores assigned to human- and AI-written proposals by both human and AI reviewers (Sec.~\ref{sec:results_rating}). Throughout, ``human reviewers'' refers to the four expert reviewers, and the two AI reviewers are Claude Opus~4.8 and ChatGPT Pro~5.5.

\subsection{Who wrote the proposal? Human or AI?}
\label{sec:results_origin}

By default, each AI reviewer was prompted, in a single pass, to score all 32 proposals on the rubric and to judge their origin. To guard against ordering or priming (earlier items biasing later judgments) effects, we also ran a two-step variant in which the proposals were rated first and their origin judged only afterwards, and we compared randomized versus sorted orderings of the proposal links; neither change significantly altered the ratings or the origin classifications (the two-step, rate-first run is labelled ``Codex~5.5~Pro'' in Table~\ref{tab:origin_id}).

\begin{table}[ht]
\centering
\begin{tabular}{|l|l|l|}
\hline
\textbf{Evaluator} & \textbf{AI-written} & \textbf{Human-written} \\
\hline
Human reviewers & 79\% & 72\% \\
\hline
Claude (Opus 4.8) & 100\% & 100\% \\
\hline
ChatGPT (5.5 Pro) & 100\% & 100\% \\
\hline
Codex (5.5 Pro, rate-first) & 100\% & 100\% \\
\hline
Claude (Sonnet 4.6) & 67\% & 100\% \\
\hline
\end{tabular}
\caption{Percentage of proposals whose author was correctly identified, for AI-written versus human-written proposals, by each evaluator. }
\label{tab:origin_id}
\end{table}

Table~\ref{tab:origin_id} reports the percentage of correct origin judgments for AI-written and human-written proposals, by evaluator. 
Four human reviewers identified AI-written proposals correctly $79\%$ of the time on average and human-written proposals $72\%$ of the time, well above chance but far from perfect, and slightly more likely to mistake a human proposal for an AI one than the reverse. Accuracy varied widely across the individual reviewers, however, from $59\%$ to $88\%$, so the human panel is far from a single clean classifier. The variation across the three AI authors was modest: when written by ChatGPT, Claude, and DeepSeek, the AI-written proposals were correctly flagged as AI by the human reviewers $84\%$, $72\%$, and $81\%$ of the time, respectively, with no single model consistently easier or harder to detect. The most capable AI reviewers, by contrast, classified every proposal correctly: Claude Opus~4.8, ChatGPT Pro~5.5, and the rate-first Codex~5.5~Pro all reached $100\%$ accuracy on all 32 proposals, for both human- and AI-written cases. We note that the proposals were presented together in a single pass, but no evaluator, human or AI, was told the ratio of human- to AI-written proposals; the perfect AI accuracy therefore does not stem from knowing the class balance. With limited proposals, however, this perfect score should not be read as evidence of reliable authorship detection in general. The near-perfect performance also appears to be a property of the frontier models only: a smaller model, Claude Sonnet~4.6, still identified every human-written proposal correctly but flagged AI-written proposals only $67\%$ of the time, below the human reviewers. 

The free-text justifications reveal what drove these judgments. The human reviewers most often flagged a proposal as AI-written when it looked ``too clean,'' had a ``very tight'' or template-like five-step structure, contained no citations, proposed unrealistic or round-number timelines, listed irrelevant or nonsensical methods and tools, or leaned on machine-learning buzzwords (e.g., ``bootstrap resampling,'' or a tacked-on large-language-model step). Conversely, they read a proposal as human when it carried concrete author--year citations or self-citations, used idiosyncratic jargon (``donuts,'' ``blobs''), adopted a focused and feasible scope, used the first person (``we''), or contained insider domain markers (e.g., a \textit{LiteBIRD} or HSC reference suggesting an author from a specific institute). These cues were applied inconsistently across the four reviewers: for the same proposal, one reviewer often read a feature as a human signature while another read it as a hint for AI, and several judgments were openly heuristic: one reviewer guessed ``AI'' simply because they ``already selected two humans.''

The two AI reviewers claimed to base their judgments on a strikingly similar set of features, but applied them consistently and without error. They marked a proposal as AI-written when it had a uniform five-phase template, polished and impersonal prose with repetitive parallel sentence structure, generic survey (large observational program) choices named without specificity, exhaustive tool lists, round month-based timelines, a boilerplate risk-mitigation step, and generic deliverables such as a ``public white paper'' or ``community code release.'' They marked a proposal as human when it cited specific papers by author and year, gave concrete parameter ranges or observational priors (e.g., a surface density of $0.2$ LBGs\,arcmin$^{-2}$, or specific laser pulse-energy ranges), used idiosyncratic jargon, or showed cautious, conditional scientific workflow (validating on simulations before real data; beginning from a Helmholtz decomposition). Notably, Claude Opus~4.8 and ChatGPT Pro~5.5 agreed with each other to a remarkable degree, frequently citing the same features of the same proposal, and their reasoning overlapped with some human justifications; unlike humans, they never misclassified.

\subsection{Proposal ratings \& pro-AI bias}
\label{sec:results_rating}
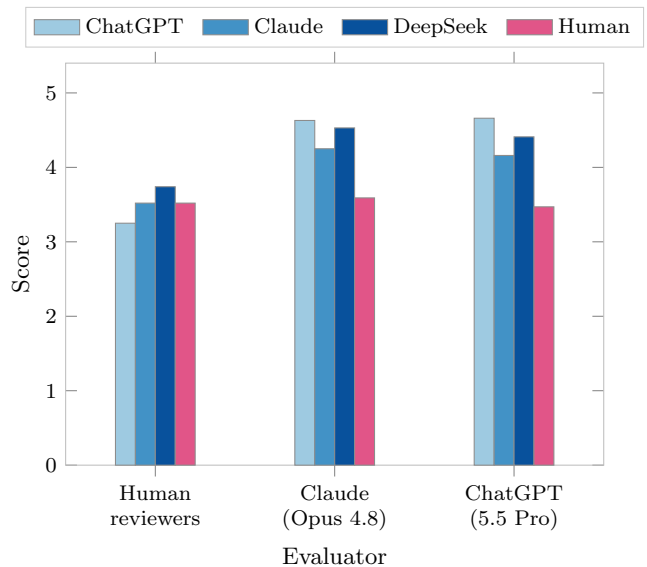
\begin{figure}[t]
\centering
\begin{tikzpicture}
\begin{axis}[
    ybar=0pt, bar width=7.5pt, width=8.7cm, height=6.9cm, area legend,
    ymin=0, ymax=5.4, ytick={0,1,2,3,4,5},
    symbolic x coords={Human reviewers, Claude Opus 4.8, ChatGPT Pro 5.5},
    xtick=data, xticklabels={Human\\reviewers, Claude\\(Opus 4.8), ChatGPT\\(5.5 Pro)},
    xticklabel style={align=center, font=\footnotesize}, enlarge x limits=0.25,
    tick label style={font=\footnotesize},
    ylabel={Score}, xlabel={Evaluator}, label style={font=\small},
    axis line style={gray!50},
    legend style={at={(0.5,1.14)}, anchor=north, legend columns=4, font=\footnotesize, draw=gray!40, /tikz/every even column/.append style={column sep=5pt}},
    legend cell align=left,
]
\addplot+[fill=aichatgpt, draw=black!45] coordinates {(Human reviewers,3.25) (Claude Opus 4.8,4.63) (ChatGPT Pro 5.5,4.66)};
\addplot+[fill=aiclaude, draw=black!45] coordinates {(Human reviewers,3.52) (Claude Opus 4.8,4.25) (ChatGPT Pro 5.5,4.16)};
\addplot+[fill=aideepseek, draw=black!45] coordinates {(Human reviewers,3.74) (Claude Opus 4.8,4.53) (ChatGPT Pro 5.5,4.41)};
\addplot+[fill=humanpink, draw=black!45] coordinates {(Human reviewers,3.52) (Claude Opus 4.8,3.59) (ChatGPT Pro 5.5,3.47)};
\legend{ChatGPT, Claude, DeepSeek, Human}
\end{axis}
\end{tikzpicture}
\caption{Mean total proposal score (averaged over the four rubric aspects), grouped by evaluator (human reviewers, Claude Opus~4.8, ChatGPT Pro~5.5). Within each group, the bars show the four proposal authors, with the three AI authors in shades of blue and the human author in pink. Human reviewers score all authors similarly ($\sim\!3.5$), whereas both AI reviewers score the AI-written proposals roughly one point above the human-written ones. Means and standard deviations are listed in Table~\ref{tab:ratings}.}
\label{fig:ratings}
\end{figure}

Fig.~\ref{fig:ratings} shows the mean total score (averaged over the four rubric aspects), grouped by evaluator, with the four proposal authors shown side by side within each group. In this and all subsequent plots, the reported standard deviations (Tables~\ref{tab:ratings} and~\ref{tab:ratings_aspects}) are taken across the eight proposals in each evaluator--author category (one per project).

The human reviewers rate all authors into a narrow range ($3.25$--$3.74$ out of $5$): DeepSeek scored highest ($3.74 \pm 0.52$), ChatGPT lowest ($3.25 \pm 0.47$), and, critically, the human-written proposals ($3.52 \pm 0.59$) in the middle, essentially the same as the AI-written average ($3.51$). By the human reviewers' own scoring, AI-written proposals were therefore no worse, and in DeepSeek's case marginally better, than the expert-written ones. We note, however, that the four human reviewers often disagreed with one another, so the human reviewers' mean is a relatively noisy measure.

\begin{table}[t]
\centering
\begin{tabular}{lccc}
\toprule
\textbf{Author} & \textbf{Human rev.} & \textbf{Claude 4.8} & \textbf{ChatGPT 5.5} \\
\midrule
ChatGPT  & $3.25\pm0.47$ & $4.63\pm0.52$ & $4.66\pm0.39$ \\
Claude   & $3.52\pm0.43$ & $4.25\pm0.54$ & $4.16\pm0.42$ \\
DeepSeek & $3.74\pm0.52$ & $4.53\pm0.25$ & $4.41\pm0.39$ \\
Human    & $3.52\pm0.59$ & $3.59\pm0.61$ & $3.47\pm0.50$ \\
\bottomrule
\end{tabular}
\caption{Mean total proposal score $\pm$ standard deviation across the eight projects, for each proposal author and evaluator. The mean scores are shown in Fig.~\ref{fig:ratings}.}
\label{tab:ratings}
\end{table}
The AI reviewers, in contrast, rated the AI-written proposals far higher: Claude Opus~4.8 gave the AI authors an average of $4.47$ versus $3.59$ for the human authors, and ChatGPT Pro~5.5 gave $4.41$ versus $3.47$, roughly one full point in favor of the AI-written proposals. The same gap appears in two further tests not shown in Fig.~\ref{fig:ratings}: the rate-first Codex~5.5~Pro (AI-written $4.51$ vs.\ human-written $3.50$) and Claude Sonnet~4.6 ($4.24$ vs.\ $3.72$), confirming that neither scoring the proposals before judging origin nor using a smaller model removes the effect. The human-written proposals received nearly the same score from every evaluator ($3.47$--$3.72$), so the disagreement is driven almost entirely by how the AI-written proposals are scored: humans judged them around $3.5$, while the AI reviewers judged them around $4.2$--$4.5$. The \textbf{pro-AI bias} appears across all four AI models and is absent from human evaluators.
\begin{figure*}[t]
\centering
\begin{tikzpicture}
\begin{groupplot}[
    group style={group size=2 by 2, horizontal sep=1.4cm, vertical sep=2.5cm},
    ybar=0pt, /pgf/bar width=7.5pt, width=8.6cm, height=5.9cm, area legend,
    ymin=0, ymax=5.4, ytick={0,1,2,3,4,5},
    symbolic x coords={Human reviewers, Claude Opus 4.8, ChatGPT Pro 5.5},
    xtick=data, xticklabels={Human\\reviewers, Claude\\(Opus 4.8), ChatGPT\\(5.5 Pro)},
    xticklabel style={align=center, font=\footnotesize}, enlarge x limits=0.25,
    tick label style={font=\footnotesize},
    ylabel={Score}, xlabel={Evaluator}, label style={font=\small}, title style={font=\small},
    axis line style={gray!50},
]
\nextgroupplot[title={Clarity and Structure of Research Plan}, legend style={at={(0.5,1.32)}, anchor=south, legend columns=4, font=\footnotesize, draw=gray!40, /tikz/every even column/.append style={column sep=5pt}}, legend cell align=left]
\addplot+[fill=aichatgpt, draw=black!45] coordinates {(Human reviewers,3.44) (Claude Opus 4.8,4.88) (ChatGPT Pro 5.5,5.0)};
\addplot+[fill=aiclaude, draw=black!45] coordinates {(Human reviewers,3.72) (Claude Opus 4.8,4.75) (ChatGPT Pro 5.5,4.88)};
\addplot+[fill=aideepseek, draw=black!45] coordinates {(Human reviewers,4.16) (Claude Opus 4.8,5.0) (ChatGPT Pro 5.5,5.0)};
\addplot+[fill=humanpink, draw=black!45] coordinates {(Human reviewers,3.69) (Claude Opus 4.8,3.75) (ChatGPT Pro 5.5,3.75)};
\legend{ChatGPT, Claude, DeepSeek, Human}
\nextgroupplot[title={Appropriateness of Methods to Scientific Goal}]
\addplot+[fill=aichatgpt, draw=black!45] coordinates {(Human reviewers,3.19) (Claude Opus 4.8,4.75) (ChatGPT Pro 5.5,4.75)};
\addplot+[fill=aiclaude, draw=black!45] coordinates {(Human reviewers,3.5) (Claude Opus 4.8,4.63) (ChatGPT Pro 5.5,4.25)};
\addplot+[fill=aideepseek, draw=black!45] coordinates {(Human reviewers,3.69) (Claude Opus 4.8,4.75) (ChatGPT Pro 5.5,4.5)};
\addplot+[fill=humanpink, draw=black!45] coordinates {(Human reviewers,3.72) (Claude Opus 4.8,4.13) (ChatGPT Pro 5.5,3.75)};
\nextgroupplot[title={Resource and Tool Planning}]
\addplot+[fill=aichatgpt, draw=black!45] coordinates {(Human reviewers,3.31) (Claude Opus 4.8,4.5) (ChatGPT Pro 5.5,4.75)};
\addplot+[fill=aiclaude, draw=black!45] coordinates {(Human reviewers,3.5) (Claude Opus 4.8,4.13) (ChatGPT Pro 5.5,4.13)};
\addplot+[fill=aideepseek, draw=black!45] coordinates {(Human reviewers,3.84) (Claude Opus 4.8,4.38) (ChatGPT Pro 5.5,4.25)};
\addplot+[fill=humanpink, draw=black!45] coordinates {(Human reviewers,3.38) (Claude Opus 4.8,3.13) (ChatGPT Pro 5.5,3.25)};
\nextgroupplot[title={Feasibility, Timeline, and Risk Awareness}]
\addplot+[fill=aichatgpt, draw=black!45] coordinates {(Human reviewers,3.06) (Claude Opus 4.8,4.38) (ChatGPT Pro 5.5,4.13)};
\addplot+[fill=aiclaude, draw=black!45] coordinates {(Human reviewers,3.38) (Claude Opus 4.8,3.5) (ChatGPT Pro 5.5,3.38)};
\addplot+[fill=aideepseek, draw=black!45] coordinates {(Human reviewers,3.28) (Claude Opus 4.8,4.0) (ChatGPT Pro 5.5,3.88)};
\addplot+[fill=humanpink, draw=black!45] coordinates {(Human reviewers,3.31) (Claude Opus 4.8,3.38) (ChatGPT Pro 5.5,3.13)};
\end{groupplot}
\end{tikzpicture}
\caption{Mean proposal score broken down by the four rubric aspects, grouped by evaluator (human reviewers, Claude Opus~4.8, ChatGPT Pro~5.5). The three AI authors are shown in shades of blue and the human author in pink. Means and standard deviations are listed in Table~\ref{tab:ratings_aspects}.}
\label{fig:ratings_aspects}
\end{figure*}
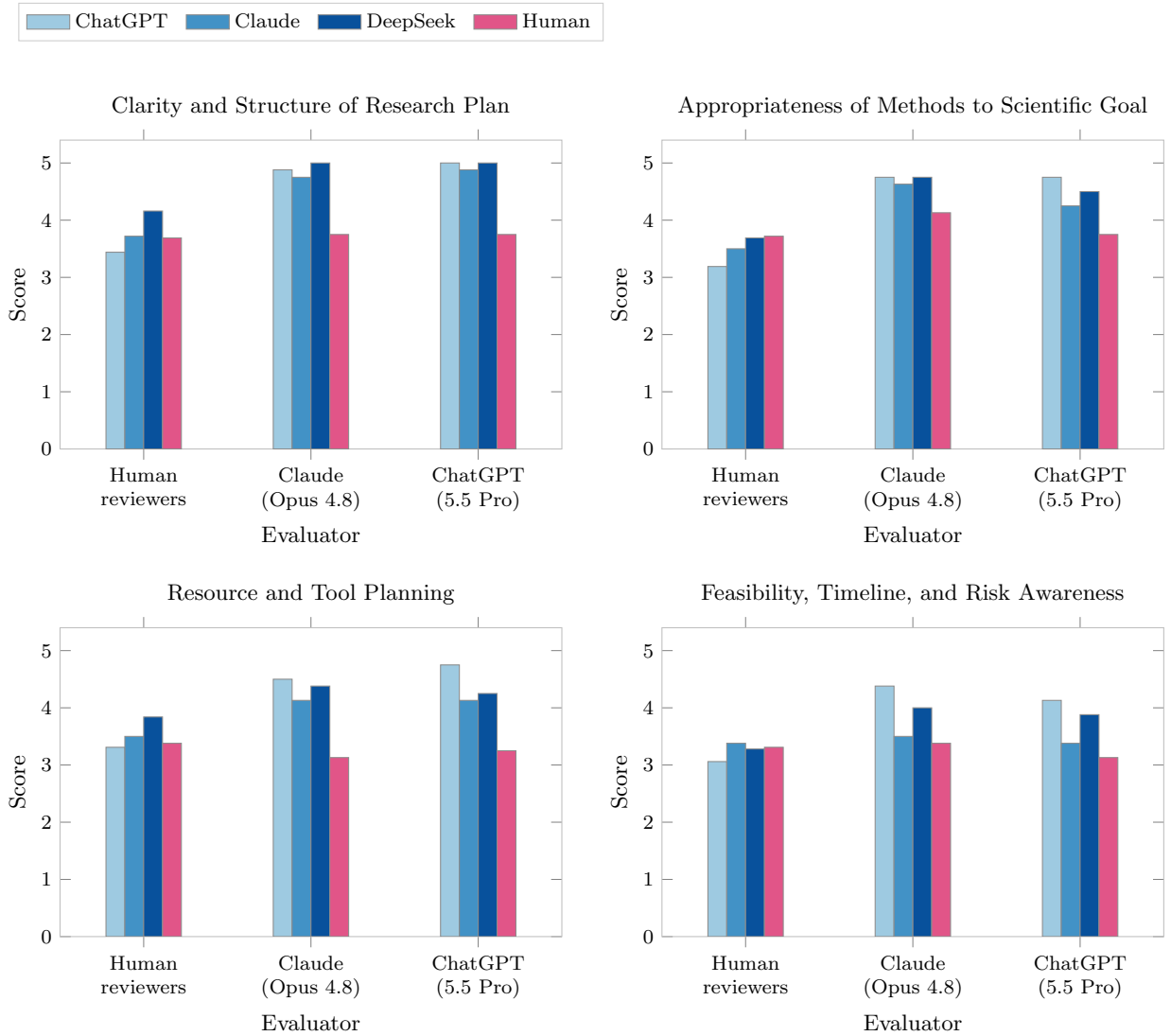

\begin{table*}[t]
\centering
\begin{tabular}{llccc}
\toprule
\textbf{Aspect} & \textbf{Author} & \textbf{Human reviewers} & \textbf{Claude (Opus 4.8)} & \textbf{ChatGPT (5.5 Pro)} \\
\midrule
Clarity and Structure      & ChatGPT  & $3.44\pm0.66$ & $4.88\pm0.35$ & $5.00\pm0.00$ \\
                           & Claude   & $3.72\pm0.41$ & $4.75\pm0.46$ & $4.88\pm0.35$ \\
                           & DeepSeek & $4.16\pm0.40$ & $5.00\pm0.00$ & $5.00\pm0.00$ \\
                           & Human    & $3.69\pm0.56$ & $3.75\pm0.46$ & $3.75\pm0.46$ \\
\midrule
Appropriateness of Methods & ChatGPT  & $3.19\pm0.46$ & $4.75\pm0.46$ & $4.75\pm0.46$ \\
                           & Claude   & $3.50\pm0.64$ & $4.63\pm0.52$ & $4.25\pm0.46$ \\
                           & DeepSeek & $3.69\pm0.78$ & $4.75\pm0.46$ & $4.50\pm0.76$ \\
                           & Human    & $3.72\pm0.49$ & $4.13\pm0.83$ & $3.75\pm0.71$ \\
\midrule
Resource and Tool Planning & ChatGPT  & $3.31\pm0.48$ & $4.50\pm0.53$ & $4.75\pm0.46$ \\
                           & Claude   & $3.50\pm0.30$ & $4.13\pm0.64$ & $4.13\pm0.35$ \\
                           & DeepSeek & $3.84\pm0.44$ & $4.38\pm0.52$ & $4.25\pm0.46$ \\
                           & Human    & $3.38\pm0.63$ & $3.13\pm0.64$ & $3.25\pm0.46$ \\
\midrule
Feasibility, Timeline, Risk & ChatGPT  & $3.06\pm0.29$ & $4.38\pm0.74$ & $4.13\pm0.64$ \\
                           & Claude   & $3.38\pm0.35$ & $3.50\pm0.53$ & $3.38\pm0.52$ \\
                           & DeepSeek & $3.28\pm0.47$ & $4.00\pm0.00$ & $3.88\pm0.35$ \\
                           & Human    & $3.31\pm0.68$ & $3.38\pm0.52$ & $3.13\pm0.35$ \\
\bottomrule
\end{tabular}
\caption{Mean proposal score $\pm$ standard deviation (across the eight projects) for each rubric aspect, proposal author, and evaluator; the data plotted in Fig.~\ref{fig:ratings_aspects}.}
\label{tab:ratings_aspects}
\end{table*}

The per-aspect breakdown in Fig.~\ref{fig:ratings_aspects} shows that these patterns are not uniform across rubric dimensions. The pro-AI bias of the AI reviewers is most pronounced for Clarity and Structure, where they rate the AI authors near the ceiling ($\sim\!4.9$--$5.0$) while holding the human author at $\sim\!3.75$, and for Resource and Tool Planning, where the human author is scored lowest of all four authors by both AI reviewers ($3.13$--$3.25$). It is weakest for Appropriateness of Methods to Scientific Goal: here the human reviewers actually place the human author at or near the top ($3.72$), and the AI-reviewer gap, while still present, is the smallest of the four aspects. Feasibility, Timeline, and Risk Awareness is consistently the weakest dimension for the AI-written proposals under every evaluator (e.g., $3.24$ from human reviewers, versus $3.46$--$3.77$ on the other three aspects), echoing the qualitative observation that AI plans tend toward unrealistic or round-number timelines and over-ambitious scope; it is also the aspect on which the human and AI proposals are most comparable. 

We find no evidence of a same-vendor preference among the AI reviewers: ChatGPT-written proposals were scored highest by both the Claude and the ChatGPT evaluators, so the pro-AI bias is a preference for AI-style writing in general rather than for an evaluator's own model family.

\subsection{Project-level variation}
\label{sec:results_projects}
\begin{figure*}[t]
\centering
\begin{tikzpicture}
\begin{groupplot}[
    group style={group size=2 by 1, horizontal sep=1.9cm},
    width=7.9cm, height=6.4cm, clip=false,
    xmin=2.3, xmax=4.2, ymin=-1.0, ymax=2.0,
    ytick={-1,-0.5,0,0.5,1,1.5,2},
    tick label style={font=\footnotesize}, label style={font=\small},
    axis line style={gray!50}, title style={font=\small\bfseries},
    ylabel={AI $-$ human score gap},
]
\nextgroupplot[title={Human panel}, xlabel={Human-written proposal score (human panel)}]
\fill[aichatgpt!30] (axis cs:2.3,0) rectangle (axis cs:4.2,2.0);
\fill[humanpink!20] (axis cs:2.3,-1.0) rectangle (axis cs:4.2,0);
\draw[gray!55, line width=0.8pt] (axis cs:2.3,0) -- (axis cs:4.2,0);
\addplot[dashed, gray!55, line width=1pt, forget plot] coordinates {(2.3,1.646) (4.2,-0.939)};
\addplot[only marks, mark=*, mark size=2.4pt, mark options={fill=aideepseek, draw=black!60}] coordinates {(3.44,0.48) (2.44,1.31) (3.25,0.54)};
\addplot[only marks, mark=*, mark size=2.4pt, mark options={fill=humanpink, draw=black!60}] coordinates {(3.75,-0.67) (3.81,-0.48) (3.5,-0.04) (4.06,-0.71) (3.94,-0.58)};
\node[font=\scriptsize, aideepseek, anchor=west] at (axis cs:2.36,0.16) {AI rated higher};
\node[font=\scriptsize, humanpink, anchor=west] at (axis cs:2.36,-0.16) {human rated higher};
\node[font=\scriptsize, anchor=north east] at (axis cs:4.15,1.9) {$r=-0.95$};
\node[font=\scriptsize, above, yshift=1pt] at (axis cs:3.44,0.48) {AGN};
\node[font=\scriptsize, above, yshift=1pt] at (axis cs:3.75,-0.67) {LBG};
\node[font=\scriptsize, above, yshift=1pt] at (axis cs:3.81,-0.48) {IA};
\node[font=\scriptsize, above, yshift=1pt] at (axis cs:3.5,-0.04) {AR};
\node[font=\scriptsize, above, yshift=1pt] at (axis cs:4.06,-0.71) {RG};
\node[font=\scriptsize, above, yshift=1pt] at (axis cs:2.44,1.31) {GW};
\node[font=\scriptsize, above, yshift=1pt] at (axis cs:3.25,0.54) {PTA};
\node[font=\scriptsize, above, yshift=1pt] at (axis cs:3.94,-0.58) {SU2};
\nextgroupplot[title={AI panel}, xlabel={Human-written proposal score (AI panel)}]
\fill[aichatgpt!30] (axis cs:2.3,0) rectangle (axis cs:4.2,2.0);
\fill[humanpink!20] (axis cs:2.3,-1.0) rectangle (axis cs:4.2,0);
\draw[gray!55, line width=0.8pt] (axis cs:2.3,0) -- (axis cs:4.2,0);
\addplot[dashed, gray!55, line width=1pt, forget plot] coordinates {(2.457,2.0) (4.2,0.180)};
\addplot[only marks, mark=*, mark size=2.4pt, mark options={fill=aideepseek, draw=black!60}] coordinates {(3.69,0.5) (3.94,0.5) (3.0,1.31) (3.44,0.98) (4.19,0.15) (3.0,1.52) (3.56,1.0) (3.75,0.73)};
\addplot[only marks, mark=*, mark size=2.4pt, mark options={fill=humanpink, draw=black!60}] coordinates {};
\node[font=\scriptsize, aideepseek, anchor=west] at (axis cs:2.36,0.16) {AI rated higher};
\node[font=\scriptsize, humanpink, anchor=west] at (axis cs:2.36,-0.16) {human rated higher};
\node[font=\scriptsize, anchor=north east] at (axis cs:4.15,1.9) {$r=-0.96$};
\node[font=\scriptsize, above, yshift=1pt] at (axis cs:3.69,0.5) {AGN};
\node[font=\scriptsize, above, yshift=1pt] at (axis cs:3.94,0.5) {LBG};
\node[font=\scriptsize, above, yshift=1pt] at (axis cs:3.0,1.31) {IA};
\node[font=\scriptsize, above, yshift=1pt] at (axis cs:3.44,0.98) {AR};
\node[font=\scriptsize, above, yshift=1pt] at (axis cs:4.19,0.15) {RG};
\node[font=\scriptsize, above, yshift=1pt] at (axis cs:3.0,1.52) {GW};
\node[font=\scriptsize, above, yshift=1pt] at (axis cs:3.56,1.0) {PTA};
\node[font=\scriptsize, above, yshift=1pt] at (axis cs:3.75,0.73) {SU2};
\end{groupplot}
\end{tikzpicture}
\caption{Project-level breakdown of the AI$-$human score gap, for the human panel (left) and the AI reviewers (right). Each point is one of the eight projects; the horizontal axis is the score that panel gave the human-written proposal, and the vertical axis is the mean AI-written score minus the human-written score. Both panels show the same strong anti-correlation with the quality of the human proposal ($r\approx-0.95$): the AI advantage is largest where the human plan was weak (e.g., GW) and smallest for the best human proposals (RG). The panels differ mainly by a vertical offset. The human panel rates the human proposal higher in five of eight projects, whereas the AI reviewers rate the AI proposals higher in all eight. So the pro-AI bias appears as an upward shift of the same trend.}
\label{fig:project_gap}
\end{figure*}
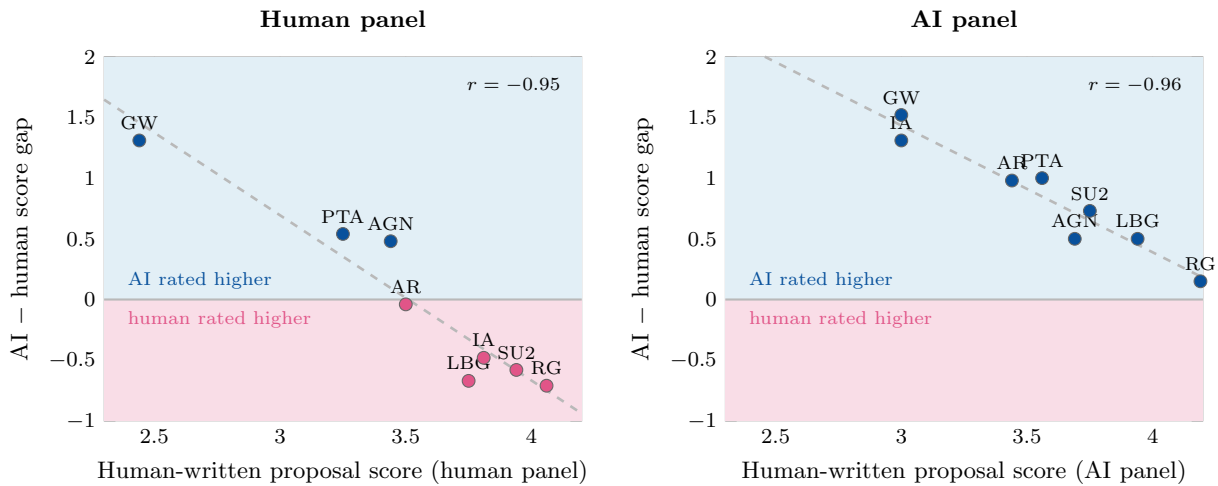

The near-parity between human- and AI-written proposals in the human panel is an aggregate that hides substantial project-to-project variation (Fig.~\ref{fig:project_gap}). The human reviewers rated the human-written proposal higher in five of the eight projects and the AI-written proposals higher in the remaining three. The size and sign of this gap are strongly anti-correlated with the quality of the human proposal itself (Pearson $r=-0.95$): where the expert produced a detailed, domain-specific plan (e.g., RG, SU2, IA), reviewers ranked it above the AI versions, whereas the AI advantage was concentrated in projects with a weak human plan, most starkly GW, whose human proposal scored only $2.4/5$. The two AI reviewers exhibit the same slope but shifted upward, preferring the AI-written proposals in all eight projects; the two panels differ mainly by a near-constant vertical offset -- the roughly one-point pro-AI bias identified above, made visual. The human--AI quality comparison and the AI-reviewer bias are therefore driven by the same handful of projects rather than being uniform across the sample, and the comparison is sensitive to the level of individual human effort on each proposal.

\section{Conclusion}
\label{sec:conclusion}

We studied human and AI performance on scientific project planning, an open-ended research-planning task with no ground-truth answer, across eight expert-conceived projects in physics, astrophysics, and cosmology. For each project, we collected one human-written and three AI-written one-page proposals and had all 32 anonymized proposals blindly evaluated for both authorship and quality by four human reviewers and two advanced AI models. Our major findings are:

\begin{enumerate}
\item On capability, at the level of a short, structured research proposal, current LLMs produce research plans comparable to those of human researchers, as rated by expert human reviewers (Figures~\ref{fig:ratings},~\ref{fig:ratings_aspects}). This demonstrates AI's potential to assist with project planning in scientific workflows.
\item Humans and AI are not equally good at telling if a proposal is written by human or AI (Table~\ref{tab:origin_id}). Human reviewers were well above chance ($\gtrsim\!70\%$) but relied on noisy and sometimes conflicting cues. The two most capable AI reviewers applied essentially the same cues as human reviewers but did so consistently and correctly, classifying all 32 proposals without error (100\%).
\item Pro-AI bias was observed in all AI evaluations, where all four AI reviewers scored AI-written proposals roughly a point higher than human-written ones (Fig.~\ref{fig:ratings}). Such bias is absent in human evaluation. This pro-AI bias is a concrete risk for any review system that incorporates LLMs: an AI reviewer may favor AI-written proposals, potentially disadvantaging human applicants and creating a feedback loop that rewards a recognizable ``AI style'' over scientific substance. The finding is consistent with previous works by \cite{Panickssery2024,Wataoka2024,Zhou2024,Shcherbiak2024}.
\footnote{Major funders have already moved to mitigate such biases: NIH prohibits the use of generative-AI tools to analyze or formulate peer-review critiques of grant applications \cite{NIH2023}, NSF bars reviewers from uploading proposal content to non-approved AI tools \cite{NSF2023}, and the ERC requires non-delegation of evaluative judgment and strict confidentiality \cite{ERC2026}; our findings supply an empirical rationale for such caution}
\item These aggregate patterns hide strong project-to-project variation (Sec.~\ref{sec:results_projects}, Fig.~\ref{fig:project_gap}): the human panel preferred the human-written proposal in five of the eight projects, and the AI--human score gap is tightly anti-correlated with the quality of the human plan ($r=-0.95$). Both the human--AI parity and the pro-AI bias are thus concentrated in a few projects rather than uniform, and the comparison is sensitive to the level of individual human effort on each proposal.
\end{enumerate}

Several caveats remain in our study. First, our sample is limited to eight projects. This reflects the substantial human effort and coordination required to advance many projects on a comparable timeline, rather than a lack of interest. Even so, the set deliberately spans a diverse range of topics and methodologies: observational, theoretical, instrumental, and numerical. Second, the AI assistants used to generate the proposals changed rapidly over the roughly one-year span of the project and were already outdated by the time of submission. Our results should therefore not be read as a benchmark of the most capable current models, but as a record of a specific point in time at which LLMs became broadly competitive with human researchers on tasks of this kind. This trend is worth documenting precisely because assembling this degree of human coordination is slow and difficult. Third, we used only a few of the most popular models (ChatGPT, Claude, and DeepSeek) and did not attempt an exhaustive or version-matched model comparison, which was not the aim of this work. Finally, we evaluated only the methodology and planning of each proposal, not the novelty of the underlying scientific idea: in our design, the ideas were all human-conceived and held fixed across the human and AI proposals. Novelty is itself a central criterion in real review, and recent work finds that LLMs can generate research ideas that expert reviewers rate as more novel than those of human researchers, albeit with somewhat weaker feasibility \cite{Si2024}; our findings are complementary to previous studies.

Future work should broaden the disciplinary and demographic scope, test full-length proposals and multi-round review, incorporate an explicit assessment of idea novelty alongside planning quality, probe whether the pro-AI bias persists when evaluators are instructed to ignore style, and extend the analysis to the other stages of the research workflow.
\begin{acknowledgments}
We thank Jingjing Shi, Qiuyue Liang, and Kenta Hotokezaka for help shaping some of the scientific projects.
JL acknowledges support from the Kavli Foundation and Google. KV acknowledges support from the Kavli Foundation. AH was supported in part by JSPS KAKENHI Grant No.~JP26K17133, the CD3 Google Seed grant, and by the World Premier International Research Center Initiative (WPI), MEXT, Japan. KK acknowledges support from the Forefront Physics and Mathematics Program to Drive Transformation (FoPM), a World-leading Innovative Graduate Study (WINGS) Program, the University of Tokyo. JG and ZL acknowledge support from the International Laboratory for Astrophysics, Neutrino and Cosmology Experiments (ILANCE). AEB is supported by the Simons Foundation. MRG acknowledges financial support from the Formaci\'on del Profesorado Universitario program of the Spanish Ministerio de Ciencia, Innovaci\'on y Universidades; from the CMB-Inflate project funded by the European Union's Horizon~2020 Research and Innovation Staff Exchange under the Marie Sk{\l}odowska-Curie grant agreement No.~101007633; and from MICIU/AEI/10.13039/501100011033 under projects PID2022-139223OB-C21 and PID2022-140670NA-I00 (also funded by FEDER, UE).
LT is supported by JSPS under KAKENHI 24K22878 and 26K17136 and by the Royal Society under ICA\textbackslash R2\textbackslash 252140.

AI usage: the AI-written proposals (Sec.~\ref{sec:proposal_writing}) were generated around mid-2025 by the AI prompters using three assistants: ChatGPT (OpenAI; GPT-4o and the o-series reasoning models, with GPT-5 following its August~2025 release), Claude (Anthropic; Claude Sonnet~4 and Opus~4), and DeepSeek (DeepSeek-V3 and R1), generally accessed through free or entry-level paid tiers. The AI evaluations (Sec.~\ref{sec:evaluation}) were performed with Claude Opus~4.8 (Anthropic) and ChatGPT Pro~5.5 (OpenAI); the additional two-step (rate-first) and cross-model tests used Codex~5.5~Pro (OpenAI) and Claude Sonnet~4.6 (Anthropic). Beyond their role as study instruments, LLMs were also used to assist with language editing, figure preparation, and formatting of this manuscript; all scientific content, analyses, and conclusions are the authors' own.
\end{acknowledgments}
\section*{Author affiliations}
{\footnotesize \raggedright \noindent
$^{1}$Center for Data-Driven Discovery, Kavli IPMU (WPI), UTIAS, The University of Tokyo, Kashiwa, Chiba 277-8583, Japan \\
$^{2}$Kavli IPMU (WPI), UTIAS, The University of Tokyo, 5-1-5 Kashiwanoha, Kashiwa, Chiba 277-8583, Japan \\
$^{3}$Department of Astrophysical Sciences, Princeton University, Peyton Hall, Princeton, NJ 08544, USA \\
$^{4}$Department of Physics, The University of Tokyo, 7-3-1 Hongo, Bunkyo-ku, Tokyo 113-0033, Japan \\
$^{5}$Center for Computational Astrophysics, Flatiron Institute, 162 5th Avenue, New York, NY 10010, USA \\
$^{6}$Department of Physics, University of Oxford, Denys Wilkinson Building, Keble Road, Oxford OX1 3RH, United Kingdom \\
$^{7}$Research Center for the Early Universe, The University of Tokyo, Bunkyo-ku, Tokyo 113-0033, Japan \\
$^{8}$\'Ecole polytechnique, Institut polytechnique de Paris, Palaiseau, France \\
$^{9}$Advanced Energy, Graduate School of Frontier Sciences, The University of Tokyo, Kashiwa, Chiba 277-8561, Japan \\
$^{10}$Department of Astronomy, University of Texas at Austin, Austin, Texas, USA \\
$^{11}$Department of Physics, Graduate School of Science, The University of Tokyo, Bunkyo-ku, Tokyo 113-0033, Japan \\
$^{12}$Instituto de F\'isica de Cantabria (IFCA, CSIC--UC), Avenida los Castros s/n, 39005 Santander, Spain \\
$^{13}$Departamento de F\'isica Moderna, Universidad de Cantabria, Avenida los Castros s/n, E-39005 Santander, Spain \\
$^{14}$Institute for Cosmic Ray Research, The University of Tokyo, 5-1-5 Kashiwa-no-Ha, Kashiwa, Chiba 277-8582, Japan \\
$^{15}$Department of Astronomy, University of Science and Technology of China, Hefei, Anhui 230026, People's Republic of China \\
$^{16}$School of Astronomy and Space Sciences, University of Science and Technology of China, Hefei, Anhui 230026, People's Republic of China \\
}
\bibliography{refs}{}

@article{Hezaveh2017,
  author  = {Hezaveh, Yashar D. and Perreault Levasseur, Laurence and Marshall, Philip J.},
  title   = {Fast automated analysis of strong gravitational lenses with convolutional neural networks},
  journal = {Nature},
  volume  = {548},
  number  = {7669},
  pages   = {555--557},
  year    = {2017},
  doi     = {10.1038/nature23463}
}

@article{VillaescusaNavarro2021,
  author  = {Villaescusa-Navarro, Francisco and Angl{\'e}s-Alc{\'a}zar, Daniel and Genel, Shy and Spergel, David N. and Somerville, Rachel S. and Dave, Romeel and Pillepich, Annalisa and Hernquist, Lars and Nelson, Dylan and Torrey, Paul and others},
  title   = {The {CAMELS} Project: Cosmology and Astrophysics with Machine-learning Simulations},
  journal = {The Astrophysical Journal},
  volume  = {915},
  number  = {1},
  pages   = {71},
  year    = {2021},
  doi     = {10.3847/1538-4357/abf7ba}
}

@inproceedings{Nguyen2023,
  author    = {Nguyen, Tuan Dung and Ting, Yuan-Sen and Ciuc{\u{a}}, Ioana and O'Neill, Charlie and Sun, Ze-Chang and Jab{\l}o{\'n}ska, Maja and Kruk, Sandor and Perkowski, Ernest and Miller, Jack and Li, Jason and others},
  title     = {{AstroLLaMA}: Towards Specialized Foundation Models in Astronomy},
  booktitle = {Proceedings of the Second Workshop on Information Extraction from Scientific Publications (WIESP), IJCNLP-AACL 2023},
  pages     = {49--55},
  year      = {2023},
  note      = {arXiv:2309.06126}
}

@inproceedings{Si2024,
  author    = {Si, Chenglei and Yang, Diyi and Hashimoto, Tatsunori},
  title     = {Can {LLMs} Generate Novel Research Ideas? A Large-Scale Human Study with 100+ {NLP} Researchers},
  booktitle = {International Conference on Learning Representations (ICLR)},
  year      = {2025},
  note      = {arXiv:2409.04109}
}

@article{Lu2024,
  author  = {Lu, Chris and Lu, Cong and Lange, Robert Tjarko and Foerster, Jakob and Clune, Jeff and Ha, David},
  title   = {The {AI} Scientist: Towards Fully Automated Open-Ended Scientific Discovery},
  journal = {arXiv e-prints},
  year    = {2024},
  note    = {arXiv:2408.06292}
}

@article{Liang2024,
  author  = {Liang, Weixin and Zhang, Yuhui and Cao, Hancheng and Wang, Binglu and Ding, Daisy Yi and Yang, Xinyu and Vodrahalli, Kailas and He, Siyu and Smith, Daniel Scott and Yin, Yian and McFarland, Daniel A. and Zou, James},
  title   = {Can Large Language Models Provide Useful Feedback on Research Papers? A Large-Scale Empirical Analysis},
  journal = {NEJM AI},
  volume  = {1},
  number  = {8},
  year    = {2024},
  doi     = {10.1056/AIoa2400196},
  note    = {arXiv:2310.01783}
}

@inproceedings{Mitchell2023,
  author    = {Mitchell, Eric and Lee, Yoonho and Khazatsky, Alexander and Manning, Christopher D. and Finn, Chelsea},
  title     = {{DetectGPT}: Zero-Shot Machine-Generated Text Detection using Probability Curvature},
  booktitle = {Proceedings of the 40th International Conference on Machine Learning (ICML)},
  series    = {PMLR},
  volume    = {202},
  pages     = {24950--24962},
  year      = {2023},
  note      = {arXiv:2301.11305}
}

@article{Gao2023,
  author  = {Gao, Catherine A. and Howard, Frederick M. and Markov, Nikolay S. and Dyer, Emma C. and Ramesh, Siddhi and Luo, Yuan and Pearson, Alexander T.},
  title   = {Comparing scientific abstracts generated by {ChatGPT} to real abstracts with detectors and blinded human reviewers},
  journal = {npj Digital Medicine},
  volume  = {6},
  number  = {1},
  pages   = {75},
  year    = {2023},
  doi     = {10.1038/s41746-023-00819-6}
}

@article{Shcherbiak2024,
  author  = {Shcherbiak, Anna and Habibnia, Hooman and B{\"o}hm, Robert and Fiedler, Susann},
  title   = {Evaluating science: A comparison of human and {AI} reviewers},
  journal = {Judgment and Decision Making},
  volume  = {19},
  pages   = {e21},
  year    = {2024},
  doi     = {10.1017/jdm.2024.24}
}

@inproceedings{Panickssery2024,
  author    = {Panickssery, Arjun and Bowman, Samuel R. and Feng, Shi},
  title     = {{LLM} Evaluators Recognize and Favor Their Own Generations},
  booktitle = {Advances in Neural Information Processing Systems (NeurIPS)},
  volume    = {37},
  year      = {2024},
  note      = {arXiv:2404.13076}
}

@article{Wataoka2024,
  author  = {Wataoka, Koki and Takahashi, Tsubasa and Ri, Ryokan},
  title   = {Self-Preference Bias in {LLM}-as-a-Judge},
  journal = {arXiv e-prints},
  year    = {2024},
  note    = {arXiv:2410.21819. Presented at the NeurIPS 2024 Safe Generative AI Workshop}
}

@inproceedings{Zhou2024,
  author    = {Zhou, Hongli and Huang, Hui and Long, Yunfei and Xu, Bing and Zhu, Conghui and Cao, Hailong and Yang, Muyun and Zhao, Tiejun},
  title     = {Mitigating the Bias of Large Language Model Evaluation},
  booktitle = {Proceedings of the 23rd Chinese National Conference on Computational Linguistics (CCL)},
  pages     = {1310--1319},
  year      = {2024},
  note      = {arXiv:2409.16788}
}

@article{Sadasivan2023,
  author  = {Sadasivan, Vinu Sankar and Kumar, Aounon and Balasubramanian, Sriram and Wang, Wenxiao and Feizi, Soheil},
  title   = {Can {AI}-Generated Text be Reliably Detected?},
  journal = {arXiv e-prints},
  year    = {2023},
  note    = {arXiv:2303.11156}
}

@article{Ren2025,
  author  = {Ren, Jing and Wang, Weiqi},
  title   = {Assisting Research Proposal Writing with Large Language Models: Evaluation and Refinement},
  journal = {arXiv e-prints},
  year    = {2025},
  note    = {arXiv:2509.09709}
}

@article{Sandstrom2026,
  author  = {Sandstr{\"o}m, Ulf and Thelwall, Mike},
  title   = {Can Large Language Models Evaluate Grant Proposal Quality? Revisiting the Wenner{\aa}s and Wold Peer Review Data},
  journal = {arXiv e-prints},
  year    = {2026},
  note    = {arXiv:2603.14565}
}

@article{Thorne2026,
  author  = {Thorne, William and James, Joseph and Wang, Yang},
  title   = {Evaluating {LLM}-Based Grant Proposal Review via Structured Perturbations},
  journal = {arXiv e-prints},
  year    = {2026},
  note    = {arXiv:2603.08281}
}

@article{Sikimic2025,
  author  = {Sikimi{\'c}, Vlasta},
  title   = {Fair or flawed? Rethinking grant review with generative {AI}},
  journal = {Synthese},
  volume  = {206},
  pages   = {282},
  year    = {2025},
  doi     = {10.1007/s11229-025-05366-z}
}

@misc{denario,
      title={The Denario project: Deep knowledge AI agents for scientific discovery}, 
      author={Francisco Villaescusa-Navarro and Boris Bolliet and Pablo Villanueva-Domingo and Adrian E. Bayer and Aidan Acquah and Chetana Amancharla and Almog Barzilay-Siegal and Pablo Bermejo and Camille Bilodeau and Pablo Cárdenas Ramírez and Miles Cranmer and Urbano L. França and ChangHoon Hahn and Yan-Fei Jiang and Raul Jimenez and Jun-Young Lee and Antonio Lerario and Osman Mamun and Thomas Meier and Anupam A. Ojha and Pavlos Protopapas and Shimanto Roy and David N. Spergel and Pedro Tarancón-Álvarez and Ujjwal Tiwari and Matteo Viel and Digvijay Wadekar and Chi Wang and Bonny Y. Wang and Licong Xu and Yossi Yovel and Shuwen Yue and Wen-Han Zhou and Qiyao Zhu and Jiajun Zou and Íñigo Zubeldia},
      year={2025},
      eprint={2510.26887},
      archivePrefix={arXiv},
      primaryClass={cs.AI},
      url={https://arxiv.org/abs/2510.26887}, 
}

@misc{NIH2023,
  author       = {{National Institutes of Health}},
  title        = {The Use of Generative Artificial Intelligence Technologies is Prohibited for the {NIH} Peer Review Process},
  howpublished = {NIH Guide Notice NOT-OD-23-149},
  year         = {2023},
  note         = {\url{https://grants.nih.gov/grants/guide/notice-files/NOT-OD-23-149.html}}
}

@misc{NSF2023,
  author       = {{National Science Foundation}},
  title        = {Notice to the Research Community: Use of Generative Artificial Intelligence Technology in the {NSF} Merit Review Process},
  year         = {2023},
  note         = {\url{https://www.nsf.gov/news/notice-to-the-research-community-on-ai}}
}

@misc{ERC2026,
  author       = {{European Research Council}},
  title        = {The Use of {AI} in Grant Proposal Evaluation},
  year         = {2026},
  note         = {\url{https://erc.europa.eu/system/files/2026-03/Use-AI-grant-proposal-evaluation.pdf}. See also \url{https://erc.europa.eu/news-events/news/erc-clarifies-limits-ai-use-grant-evaluation}}
}

@misc{Hell:2026dqx,
    author = "Hell, Anamaria and Thiele, Leander",
    title = "{LLMs with in-context learning for Algorithmic Theoretical Physics}",
    eprint = "2605.08212",
    archivePrefix = "arXiv",
    primaryClass = "cs.LG",
    reportNumber = "IPMU26-0020",
    month = "5",
    year = "2026"
}

@misc{Paper1,
    author = "Hell, Anamaria and Vovk,  Kateryna and Krishnaraj, Veena and Liu, Jia and Aizawa, Kosuke and Bayer, Adrian E. and Blot, Linda and Cowell, Jessica and Garg, Suyog and Gr\'ee, Jonathan and Horowitz, Ben and Ichikawa, Masaya and Iemoto, Kanyuni and Kondo, Keigo and Lorsin, Zacharie and McCarthy, Kevin and Robinson, Jamie and Ruiz-Granda, Miguel and Thiele, Leander and Vovk, Ievgen and Zhou, Mingshen",
    title = "{AI’s Capability in Assisting Scientific Research in Physics, Astrophysics, and
Cosmology I: Literature Review}",
    eprint = "2607.xxxx",
    archivePrefix = "arXiv",
}
\bibliographystyle{apsrev4-2}
\appendix
\section{Background and goals of the eight research projects}
\label{app:project_background}
The following project titles, backgrounds, and goals were written by the human experts without any AI assistance, and were the common input provided to the human planners and the AI prompters for each project (Sec.~\ref{sec:projects}).
\subsection{AGN  --  MaNGA: AGN Duty Cycle}
\textit{Background:} The time a galaxy spends in the AGN phase, from both general arguments and ensemble studies such as quasar clustering and black-hole mass-function studies and He\,\textsc{ii} proximity-zone analysis, is suggested to last $\sim\!10^{6}$--$10^{9}$ yr. Ionization studies of AGN host galaxies and their surroundings indicate that active nuclei can switch on/off on a $\sim\!100$ kyr timescale -- multiple times during their lifetime -- also finding support in numerical simulations, indicating that the AGN luminosity may fluctuate by orders of magnitude over its lifetime. For a galaxy size of $R\sim10$ kpc it takes $t=R/c\sim30$ kyr for a change in AGN UV luminosity to propagate over the entire galactic disk. We propose to perform a systematic search for fading or, more generally, variable AGN in the most recent MaNGA survey DR17. \\
\textit{Goal:} To assess the total fraction of fading/brightening AGN among the sample, indicative of the duty-cycle fraction these objects span in the transitional phase, directly comparable to the predictions of cosmological simulations. We further aim to characterize the AGN UV luminosity evolution on kyr timescales and compare it, where possible, to that derived from the resolved longitudinal brightness profiles of X-ray jets, potentially yielding a first-of-its-kind link between AGN accretion rate and high-energy particle acceleration in AGN jets.

\subsection{LBG  --  The galaxy--dark matter halo connection of Lyman-break galaxies}
\textit{Background:} A Lyman-break galaxy (LBG) is a galaxy whose broadband photometry shows a ``drop-out'' in the bluest bands as features in its spectrum move from blue to red through the filter set due to cosmic expansion; the reduction in flux blueward of the Lyman-$\alpha$ and Lyman-limit frequencies is caused by scattering and absorption of UV photons by intervening neutral hydrogen. This feature is a robust way to identify galaxies at $z>2$ in photometric surveys, which can then be followed up spectroscopically. Current and future surveys, such as the `Ōnohi`ula Prime Focus Spectrograph Galaxy Evolution Survey (PFS:GE), will measure redshifts for thousands of LBGs at $2.0<z<4.0$ to study their properties and how they trace large-scale structure (LSS). \\
\textit{Goal:} In preparation for the measurement and modeling of the LBG galaxy--galaxy two-point correlation function (2PCF), we will conduct a theoretical investigation into the potential progenitors of the PFS:GE LBG target sample using the Uchuu--UniverseMachine mock galaxy catalog, to better understand how they trace the LSS via galaxy bias and the halo occupation distribution (HOD) as a function of redshift. Of particular interest is any sign of galaxy assembly bias or other selection-function nuances that could lead to incorrect inferences of the galaxy bias ($b_g$) and growth rate of LSS ($f\sigma_8$).
\subsection{IA  --  Intrinsic alignments in varying environments}
\textit{Background:} Weak-lensing surveys are one of the most powerful probes in cosmology; however, the intrinsic alignment (IA) of galaxies contaminates the signal. Many studies have therefore investigated the characteristics of IA in order to eliminate it from the data. So far, researchers believe IA is related to galaxy color and luminosity; in addition, we suspect it is related to the large-scale environment, such as the matter distribution. \\
\textit{Goal:} We will evaluate the additional dependence of IA on the large-scale matter density field and estimate the impact on cosmological analysis when such a dependence is neglected.
\subsection{AR  --  Prediction of debris emergence on laser-ablated sub-wavelength shapes}
\textit{Background:} We have been developing methods to fabricate sub-wavelength structures (SWS) for anti-reflective coating in the millimeter-wave region on hard materials such as ceramics, using ultra-short-pulse laser ablation, which is crucial for machining materials with relatively wide energy band gaps. For efficient fabrication of SWS with a higher ablation rate it is necessary to inject lasers with as high a power as possible; however, at the same time we observe redeposition of ablated debris, particularly at higher energies. Since the physics behind the emergence of debris is mostly unknown, their shape is uncontrollable, and the quantitative conditions under which they appear are ambiguous. \\
\textit{Goal:} To predict the shape expected to be fabricated for a given set of laser-scanning parameters (pulse frequency, pulse energy, spacing of scanning lines, etc.), we first accumulate data on fabricated shapes across different parameter sets and make plots of the results including shape information such as depth and the existence of debris. Our goal is to extrapolate to arbitrary parameters using Gaussian-process regression (and, where useful, large language models), comparing predicted data points against real laser-machining data.
\subsection{RG  --  Radio Galaxies with HalfDome}
\textit{Background:} At low CMB frequencies (around 100 GHz), high-energy radio galaxies act as bright point-source contaminants to CMB maps. The locations of these galaxies are likely correlated with features in the underlying large-scale structure as well as with galaxy properties (e.g., the CIB, radio continuum, X-ray). \\
\textit{Goal:} Add realistic radio luminosities to galaxies in N-body simulations that are relevant for CMB foreground-contamination modeling and correlated with the mass of the underlying haloes, incorporating as much accurate physics as possible (such as the synchrotron-spectrum behavior of different radio-galaxy types). The HalfDome simulations aim to create realistic correlated foregrounds across different wavelengths on N-body simulations.

\subsection{GW  --  Environment of gravitational-wave black hole binaries with weak-lensing maps}
\textit{Background:} The first detection of a gravitational wave (GW) by LIGO opened a new era of multi-messenger astronomy, and around 300 GW events from binary black hole (BBH) mergers have now been observed. However, the origins of these binary black holes and the environments they reside in remain unknown. Combining gravitational waves with cosmological data will provide rich information on the origin and evolution of BBHs. \\
\textit{Goal:} This project aims to unveil the environment of BBHs. We combine publicly available gravitational-wave localization data with cosmological data, such as weak-lensing maps and galaxy-overdensity maps, to investigate the correlation between the BBH distribution and the matter distribution in the universe. The results will be used to constrain the astrophysical origins of BBHs.

\subsection{PTA  --  Forecasting pulsar timing array sensitivity to deviations from general relativity}
\textit{Background:} The Pulsar Timing Array (PTA) is a measurement method relying on the observation of pulsars, fast-rotating neutron stars with well-known timing models. By measuring slight perturbations in the times of arrival (ToAs) of each pulse, computing residuals, and plotting the pulsar-pair correlations as a function of angular separation, one obtains the overlap reduction function (ORF), known in GR as the Hellings--Downs curve. This method allowed, in 2023, evidence for a stochastic gravitational-wave background to be reported by several collaborations (EPTA, NANOGrav, CPTA, \ldots). In modified gravity, additional polarization modes and a modified dispersion relation lead to a modified ORF. \\
\textit{Goal:} In our work we neglect the scalar and vector polarization modes and focus on a modified dispersion relation (which modifies the phase and group velocities). Approximating the gravitational-wave background as plane waves interfering with each other, the relevant quantity to study is the phase velocity, which we do not constrain a priori (parameter space $[0,+\infty)$). Our goal is to forecast the measurement precision required to distinguish, at $1,2,3$ to $5\sigma$ confidence, a $20\%$, $10\%$ to $1\%$ deviation from GR (in terms of phase-velocity deviation), and in how many years that precision will be achievable.

\subsection{SU2  --  Massive Yang--Mills theory}
\textit{Background:} When exploring mechanisms that drive inflation, non-Abelian gauge fields -- such as SU(2) Yang--Mills fields -- have been proposed as alternatives to scalar fields. For instance, introducing a Chern--Simons coupling between a pseudo-scalar field and a non-Abelian gauge field can lead to slow-roll inflation, as in Chromo-Natural Inflation. Moreover, higher-order gauge-field corrections, such as $(F\tilde{F})^2$ terms, can emerge in effective field theories and play a significant role in early-universe dynamics, potentially enabling inflationary scenarios. \\
\textit{Goal:} The goal of this project is to investigate mechanisms by which vector fields can drive cosmic inflation. We focus on Proca and Yang--Mills SU(2) theories and study how to modify them so that their background solution leads to accelerated expansion on cosmological scales, with particular emphasis on the effect of adding a mass term in both theories compared to the massless case.

\section{Standardized prompts and instructions}
\label{app:project_plan_prompt}

Both human planners and AI prompters were asked to produce a one-page proposal, in a formal scientific tone appropriate for an academic review panel, organized under four headings:

\begin{enumerate}[noitemsep]
\item Project Title, copied from the title provided;
\item Background, a one-sentence summary;
\item Goal, a one-sentence summary; and
\item Methodology, broken into no more than five major steps or phases (e.g., data preparation, modeling, analysis), each with an approximate completion time, in under 300 words.
\end{enumerate}

All parties were told that the proposal would be evaluated on four criteria: clarity and structure of the research plan; appropriateness of methods to the scientific goal; resource and tool planning; and feasibility, timeline, and risk awareness (Table~\ref{tab:project_plan_rubric}).
The AI prompters prompted each model with the following fixed template:
\begin{quote}
\small
You are an expert physics researcher writing a 1-page proposal for a new research project. The audience is an academic review panel.\\[2pt]
The project title: [provided by human]\\
Background: [provided by human]\\
Goal: [provided by human]\\[2pt]
Write the proposal using clear, concise academic language and organize it under the following headings and instructions:
1.~Project Title: copy from the title provided;
2.~Background: one sentence summary;
3.~Goal: one sentence summary;
4.~Methodology: Break down the work into major steps or phases (e.g., data preparation, modeling, analysis). Include theoretical, computational, or experimental techniques if relevant. Include the approximate time for each step. No more than 5 steps and keep it under 300 words.\\[2pt]
Use formal scientific tone appropriate for a grant or academic setting. Your proposal will be evaluated based on the following criteria: 1.~Clarity and structure of research plan, 2.~Appropriateness of methods to scientific goal, 3.~Resource and tool planning, 4.~Feasibility, timeline, and risk awareness.
\end{quote}

To remove superficial stylistic tells, each human-written proposal was passed through ChatGPT with the following instruction: ``Correct typo and grammar mistakes, with minimal change to the content: [text].''

\section{Sample proposals for the SU(2) project}
\label{app:su2_samples}
To illustrate the material that reviewers saw, we reproduce the four proposals for the SU(2) project (one human, three AI), all generated from the identical title, background, and goal in Appendix~\ref{app:project_background}. They are lightly reformatted for typesetting but otherwise unedited.
\begin{proposalbox}{Human}
\textit{Title:} Massive Yang--Mills theory (SU2). \\[6pt]
\textit{Background:} In the context of early-universe cosmology, several models have proposed that vector fields -- either through direct coupling to the inflaton scalar field or via higher-order terms such as $F^4$ -- can act as viable sources of inflation. \\[6pt]
\textit{Goal:} This internship tackles vector-field theories as candidates driving cosmic inflation, starting with Proca theory and extending it to non-Abelian massive SU(2) Yang--Mills theory, with a particular focus on their coupling to gravity and the effect of the addition of the mass. \\[6pt]
\textit{Methodology:}
\begin{enumerate}
\item \textit{Theoretical foundations \& Helmholtz decomposition (2 weeks):} Review Proca theory in curved spacetime; use the Helmholtz decomposition to separate the vector field into longitudinal and transverse modes; derive the equations of motion and identify constraint equations; compare with the massless (Maxwell) case to isolate physical degrees of freedom.
\item \textit{Energy--momentum tensor \& inflation criteria in an FRW background (2--3 weeks):} Apply the theory to a homogeneous, isotropic FRW background; derive the energy--momentum tensor from the Proca Lagrangian and compute the energy density and pressure; analyze whether the configuration can drive inflation (i.e., satisfies $\rho+3p<0$).
\item \textit{Extension to massive SU(2) Yang--Mills theory (2--3 weeks):} Generalize to the non-Abelian SU(2) case with mass; derive the equations of motion and compute energy density and pressure in FRW spacetime; compare with Proca theory and evaluate under what conditions the SU(2) configuration leads to inflationary dynamics.
\item \textit{Computational implementation \& numerical solutions (3--4 weeks):} Implement the background field equations numerically in Python or Mathematica; study the time evolution of the scale factor, energy density, and field components for various initial conditions and coupling parameters; identify regions in parameter space favorable for inflation.
\item \textit{Perturbation analysis \& stability (3--4 weeks):} Linearize the field equations around the inflationary background and classify perturbations (scalar, vector, tensor); study their evolution to assess the stability of the background solution; analyze whether instabilities or anisotropies arise in either the Proca or SU(2) case.
\end{enumerate}
\end{proposalbox}
\begin{proposalbox}{ChatGPT}
\textit{Title:} Massive Yang--Mills theory (SU2). \\[6pt]
\textit{Background:} Non-Abelian SU(2) gauge fields augmented by Chern--Simons and higher-order $F\tilde{F}$ operators offer a viable alternative to scalar inflation models, but the cosmological role of an explicit mass term remains unexplored. \\[6pt]
\textit{Goal:} Quantitatively determine how adding a mass term to Proca and SU(2) Yang--Mills fields alters background dynamics and perturbative stability, identifying parameter regimes that support sustained, isotropic inflation. \\[6pt]
\textit{Methodology:}
\begin{enumerate}
\item \textit{Model formulation (4 months):} Derive covariant Lagrangians for massive Proca and SU(2) sectors including $F^2$, $F\tilde{F}$, and $(F\tilde{F})^2$ corrections; impose cosmological ans\"atze; obtain Euler--Lagrange equations.
\item \textit{Background dynamics \& stability (4 months):} Solve background equations numerically across mass and coupling grids; compute slow-roll parameters, anisotropy measures, and identify fixed points; perform linear stability (tensor, vector, scalar) via the Mukhanov--Sasaki formalism, using Python/Julia ODE solvers and Mathematica for eigen-analysis.
\item \textit{Perturbation spectra \& observables (4 months):} Evolve perturbations through horizon crossing; derive power spectra, tensor chirality, and non-Gaussianity; compare with CMB--LiteBIRD sensitivities and PTA gravitational-wave forecasts.
\item \textit{Parameter estimation \& feasibility checks (3 months):} Constrain viable regions using Planck+BICEP/Keck data; propagate instrumental uncertainties to inflationary parameters; assess robustness against radiative corrections and strong-coupling bounds.
\item \textit{Risk mitigation \& dissemination (5 months):} If the mass term triggers gradient instabilities, pivot to St\"uckelberg-like UV completions; cross-check with lattice simulations; publish results in JCAP; present at COSMO-26.
\end{enumerate}
\end{proposalbox}
\begin{proposalbox}{Claude}
\textit{Title:} Massive Yang--Mills theory (SU2). \\[6pt]
\textit{Background:} This project investigates non-Abelian gauge fields as inflationary drivers, building on established frameworks like Chromo-Natural Inflation where SU(2) Yang--Mills fields coupled to pseudo-scalar fields can generate slow-roll inflation through Chern--Simons interactions and higher-order gauge corrections. \\[6pt]
\textit{Goal:} The primary objective is to systematically analyze how Proca and Yang--Mills SU(2) vector-field theories can be modified to produce cosmological inflation, with particular emphasis on comparing the dynamical effects of mass terms versus massless configurations in driving accelerated expansion. \\[6pt]
\textit{Methodology:}
\begin{enumerate}
\item \textit{Theoretical framework development (3 months):} Derive modified Einstein field equations for SU(2) Proca and Yang--Mills theories in cosmological backgrounds; establish the action formalism incorporating mass terms and analyze the stress--energy tensor components; identify key parameters controlling the transition from massless to massive regimes.
\item \textit{Background solution analysis (4 months):} Solve the coupled Einstein--Yang--Mills/Proca equations numerically (Mathematica, Python); construct a phase-space analysis to identify inflationary attractors and determine slow-roll conditions; map the parameter space where accelerated expansion is viable.
\item \textit{Perturbation theory and stability (3 months):} Perform linear perturbation analysis around inflationary backgrounds to assess stability and compute scalar/tensor power spectra; implement a gauge-invariant formalism for vector perturbations; verify that ghost instabilities are absent in the massive case.
\item \textit{Observational constraints (2 months):} Compare theoretical predictions with Planck CMB data, focusing on spectral indices, tensor-to-scalar ratio, and non-Gaussianity signatures; establish observationally viable parameter ranges and distinguish between Proca and Yang--Mills scenarios.
\item \textit{Comparative analysis and documentation (3 months):} Synthesize results comparing massive versus massless theories, identify novel inflationary mechanisms, and prepare manuscripts for publication; develop numerical codes for community use and present findings at international conferences.
\end{enumerate}
\end{proposalbox}
\begin{proposalbox}{DeepSeek}
\textit{Title:} Massive Yang--Mills theory (SU2). \\[6pt]
\textit{Background:} Non-Abelian gauge fields, notably SU(2) Yang--Mills fields coupled via mechanisms like Chern--Simons terms or through higher-order corrections such as $(F\tilde{F})^2$, present viable alternatives to scalar fields for driving cosmic inflation by enabling sustained slow-roll dynamics in the early universe. \\[6pt]
\textit{Goal:} This project aims to systematically investigate how massive vector fields, specifically within modified Proca and SU(2) Yang--Mills theories, can generate viable cosmic inflation on cosmological scales, comparing the effects of explicit mass terms against their massless counterparts on the background accelerated expansion. \\[6pt]
\textit{Methodology:}
\begin{enumerate}
\item \textit{Theoretical framework \& model specification (2 months):} Formally define the action for the massive Proca and SU(2) Yang--Mills theories, including the explicit mass term and necessary potential or coupling structures to ensure consistent isotropic background configurations; derive the background equations of motion in a flat FRW spacetime.
\item \textit{Background dynamics computation (3 months):} Implement numerical codes (Python/Mathematica, using ODE solvers like \texttt{scipy.integrate.solve\_ivp}) to solve the coupled Friedmann and gauge-field background-evolution equations; scan parameter spaces (mass scale, coupling constants, initial field values) to identify regions yielding sufficient e-folds of quasi-de Sitter expansion; quantify inflationary observables for both massive and reference massless cases.
\item \textit{Stability analysis (3 months):} Perform a comprehensive linear stability analysis of the identified inflationary solutions against isotropic perturbations, deriving and numerically solving the perturbation equations to assess the presence of ghosts, Laplacian, or gradient instabilities.
\item \textit{Perturbation spectrum \& observables (3 months):} Calculate the primordial scalar and tensor perturbation spectra; compute key observables (scalar spectral index $n_s$, tensor-to-scalar ratio $r$, non-Gaussianity $f_{\rm NL}$); compare predictions between massive Proca, massive SU(2), and their massless counterparts.
\item \textit{Synthesis, comparison \& publication (1 month):} Integrate results, evaluating the viability and distinguishing features of massive vector inflation; assess the specific role of the mass term in enabling stable inflation and shaping observational signatures; prepare findings for publication.
\end{enumerate}
\end{proposalbox}
\end{document}